\PassOptionsToPackage{table}{xcolor}
\documentclass{article}
\usepackage[T1]{fontenc}

\usepackage{iclr2027_conference,times}

\usepackage{amsmath,amssymb}
\usepackage{booktabs}
\usepackage{graphicx}\usepackage{placeins}\usepackage{wrapfig}
\usepackage{hyperref}
\usepackage{url}

\usepackage{enumitem}
\usepackage{algorithm}
\usepackage{algpseudocode}

\usepackage{booktabs}
\usepackage{multirow}
\usepackage[table]{xcolor}
\usepackage{graphicx}

\title{LegalEvolver: Agentic Legal Judgment Prediction via Rubric-Guided Self-Evolution}

\author{Xinyu Yang$^1$, Jiejun Tan$^1$, Chenlong Deng$^1$, Chenghao Zhang$^1$, Tongyu Wen$^1$, Zhicheng Dou$^1$ \\
$^1$Renmin University of China \\\texttt{\{yxygsai, dou\}@ruc.edu.cn} \\
\url{https://github.com/yxy-919/LegalEvolver}
}
\iclrfinalcopy

\begin{document}\maketitle
\lhead{Preprint}

\begin{abstract}
As LLM-based agents develop, they have been increasingly applied in the legal domain. Legal Judgment Prediction~(LJP) requires models to conduct multi-step reasoning and external legal evidence retrieval. Existing legal search agents' performance is often constrained by the quality of retrieval queries. They often formulate queries directly from the initial user request, overlooking latent legal issues and legally salient details. To address this issue, we propose \emph{LegalEvolver}, a query optimization framework specifically designed to generate legally informative queries for LJP-oriented retrieval. Sampling high-quality optimized queries and their corresponding search trajectories is challenging, as relying solely on unguided high-temperature rollouts is both inefficient and constrained by the model's initial capabilities. We therefore heuristically optimize queries as seeds, improving the efficiency of trajectory generation. Furthermore, we summarize rollout experience into rubrics and use them to guide multi-round iterative self-evolution. Iteratively performing rubric optimization and rubric-guided RL enables the model to surpass its initial performance ceiling. End-to-end performance on extensive benchmarks demonstrates the effectiveness of query optimization on LJP. Comparisons with direct SFT followed by RL demonstrate the effectiveness of rubric guidance. Evolving rubrics further outperform fixed rubrics, confirming the benefits of self-evolution.
\end{abstract}

\section{Introduction}

Legal judgment prediction (LJP) requires a model to determine the applicable law articles, charges, and prison term of a case from its fact description~\citep{DBLP:conf/acl/XuWCPWZ20}. Like a judge, the model must establish the charge before determining the sentence~\citep{DBLP:conf/emnlp/ZhongGTX0S18}. Decisions must be based on specific legal provisions, which come from statutes, judicial interpretations, and sentencing guidelines~\citep{DBLP:conf/emnlp/WuZL0LZS0K23,DBLP:journals/access/WangZHSZ24}. Because these provisions are numerous and frequently revised, parametric knowledge alone cannot capture them reliably~\citep{DBLP:conf/emnlp/DengMD24,DBLP:conf/acl/YangDD26}. Therefore, recent work turns to legal search agents, which consult these sources along with precedents during reasoning and base their predictions on the retrieved evidence~\citep{DBLP:conf/emnlp/WuZL0LZS0K23}.

However, access to retrieval alone brings modest gains~\citep{DBLP:journals/access/WangZHSZ24}. Our analysis shows that queries themselves are often the problem of failure; this diagnosis is consistent with recent work on query optimization for search agents~\citep{DBLP:conf/emnlp/LiDJZZZZD25, DBLP:conf/sigir/WenDD26}. Consider a defendant who borrows the victim's phone and then runs off with it. The query ``elements of the crime of fraud'' fails because the key point is the victim’s intent in handing over the phone, which is exactly what separates fraud from theft. A better query is like ``took a borrowed phone and fled, whether the victim disposed of property, theft versus fraud''. 
To understand what distinguishes effective queries more generally, we examine the reasoning and search trajectories of legal experts on judgment prediction. Experts rarely search for a charge or an article in isolation. Each query addresses a question still open at that point of their reasoning, and changes as the reasoning moved on. 
When deciding the charge, they search statutes and judicial interpretations; when deciding the sentence, they searched sentencing guidelines and precedents. 
In both stages, they wrote the decisive facts of the case into the query~\citep{DBLP:conf/sigir/WenDD26}.

Search agents are commonly trained with SFT followed by RL with outcome rewards~\citep{DBLP:conf/emnlp/LiXZYZYHLWL25}. This recipe struggles to inject the behaviour described above for the following reasons~\citep{DBLP:conf/icml/ChuZYTXSLL025, DBLP:conf/emnlp/ZhengAWWW25}: (1)~\textbf{Exploration stays anchored to the model's initial hypothesis}~\citep{DBLP:conf/icml/ChuZYTXSLL025}. Once the base model settles on a candidate charge, its queries keep revolving around it. Higher sampling temperature hardly varies charge under examination~\citep{DBLP:conf/nips/ChenSLSZZWPZCYZ25}. (2)~\textbf{Outcome rewards cannot tell which search went wrong}~\citep{DBLP:conf/emnlp/ZhengAWWW25}. It fails to distinguish different search failure types~\citep{DBLP:conf/emnlp/LiDJZZZZD25}. General query clarity and informativeness also miss these domain-specific legal requirements~\citep{DBLP:conf/emnlp/DengMD24,DBLP:conf/acl/LiuXYHYZW26}. (3)~\textbf{What makes a query useful changes as the agent improves}~\citep{DBLP:journals/corr/abs-2511-19399,DBLP:journals/corr/abs-2507-17746}. Once the agent routinely includes key facts, its remaining errors can depend on subtler distinctions~\citep{DBLP:conf/acl/XuWCPWZ20}. Fixed criteria fail to separate good queries from bad ones~\citep{DBLP:journals/corr/abs-2511-19399}.

The key missing ingredient is supervision for individual queries. Our core observation is that differences in judgment outcomes can provide it. When trajectories on the same case reach different judgments, the differences between their queries suggest which searches are more effective. The type of error further indicates which searches to examine. A wrong charge points to the searches for legal qualification, while a wrong sentence points to the sentencing searches. Building on this observation, we induce query criteria from within-case contrasts and assemble them into rubrics that reward each search. As the policy improves, we rebuild the rubrics from its new trajectories. The supervision thus evolves with the policy.

To address these challenges, we propose LegalEvolver, a framework for training legal search agents. It learns query rubrics from the agent's own trajectories and keeps refreshing them as the agent improves. Training consists of three consecutive stages. \textbf{(1) Diverse Trajectory Generation and Policy Initialization.} We first sample diverse reasoning and retrieval trajectories at a high temperature, then apply expert-inspired heuristics to rewrite their retrieval queries, producing additional candidate trajectories. Rejection sampling selects successful trajectories for supervised fine-tuning, equipping the agent with effective initial reasoning and retrieval behaviors. \textbf{(2) Initial Rubric Construction.} To construct query-level supervision, we contrast successful trajectories with failed trajectories for the same problem, examining how differences in query construction contribute to their divergent outcomes. We distill these comparisons into an initial set of legal query rubrics that provide explicit criteria for evaluating intermediate queries. \textbf{(3) Rubric--Policy Co-Evolution.} Building on this initialization, we jointly improve the query rubrics and the agent policy through iterative self-evolution. We combine rubric-based rewards with outcome rewards for reinforcement learning, providing feedback on both intermediate query quality and final prediction correctness. As the updated policy generates new successful trajectories, we use them to re-induce the rubrics, incorporating retrieval strategies discovered during training. The revised rubrics then guide the next round of policy optimization. Alternating these steps lets supervision evolve with the agent's capabilities, enabling improvement beyond the initial demonstrations and fixed rubric criteria.

Experiments on single- and multi-defendant legal judgment prediction show that LegalEvolver outperforms agentic retrieval-augmented baselines on most metrics, with further gains from reinforcement learning after supervised initialization.

To sum up, our contributions are threefold:
(1)~We identify query construction as a critical bottleneck for search agents in LJP.
(2)~We propose \emph{LegalEvolver}, which derives legal query rubrics from contrasting successful and failed trajectories.
(3)~We develop a self-evolution training paradigm that jointly improves agent policies and query rubrics.

\section{Related Work}
\label{sec:related_work}
\paragraph{Legal Judgment Prediction}

Legal Judgment Prediction (LJP) has been studied using neural classification models~\citep{DBLP:conf/emnlp/ZhongGTX0S18}, LLM-based legal reasoning~\citep{DBLP:conf/icail/JiangY23}, and agents that acquire external legal knowledge~\citep{DBLP:conf/acl/YangDD26}. Early work models dependencies among judgment subtasks with TopJudge~\citep{DBLP:conf/emnlp/ZhongGTX0S18}, distinguishes confusing law articles with LADAN~\citep{DBLP:conf/acl/XuWCPWZ20}, and incorporates case circumstances with NeurJudge~\citep{DBLP:conf/sigir/Yue0JWZACYW21}. With the emergence of LLMs, legal syllogism prompting and syllogistic judgment analysis explicitly structure legal reasoning~\citep{DBLP:conf/icail/JiangY23,DBLP:conf/emnlp/DengPKCWLRCR23}. ADAPT further employs an Ask--Discriminate--Predict framework to distinguish confusing charges~\citep{DBLP:conf/emnlp/DengMZD24}. External knowledge also supports judgment prediction through precedent-enhanced LLM and domain-model collaboration~\citep{DBLP:conf/emnlp/WuZL0LZS0K23} and multi-stage knowledge integration~\citep{DBLP:journals/access/WangZHSZ24}. GLARE expands confusing charge candidates and retrieves reasoning paths from similar cases for comparative reasoning~\citep{DBLP:conf/acl/YangDD26}. Query formulation is itself relevant to legal retrieval: knowledge-guided case reformulation has been studied for interpretable case matching~\citep{DBLP:conf/emnlp/DengMD24}. We focus on learning to formulate intermediate retrieval queries within an LJP agent, using case facts to seek the legal knowledge needed for judgment reasoning.

\paragraph{Rubric-Based Supervision for LLMs}
Rubric-based supervision expresses response quality through explicit criteria, supporting both model evaluation~\citep{DBLP:conf/emnlp/LiuIXWXZ23} and policy optimization~\citep{DBLP:journals/corr/abs-2507-17746}. G-Eval combines specified evaluation criteria with generated evaluation steps and structured scoring~\citep{DBLP:conf/emnlp/LiuIXWXZ23}. AutoCalibrate induces scoring criteria from human-labeled examples, then selects and refines them to improve agreement with human judgments~\citep{DBLP:conf/coling/LiuYHZHWDSZ24}. For contrastive construction, OpenRubrics derives explicit constraints and implicit quality principles by comparing preferred and rejected responses, producing criteria that distinguish their quality~\citep{DBLP:conf/acl/LiuXYHYZW26}. Rubrics as Rewards aggregates instance-specific rubric feedback into rewards for on-policy RL beyond verifiable domains~\citep{DBLP:journals/corr/abs-2507-17746}. DR Tulu further maintains evolving rubrics that adapt with the policy to provide discriminative feedback during deep-research training~\citep{DBLP:journals/corr/abs-2511-19399}. These works motivate preference-based criterion induction~\citep{DBLP:conf/acl/LiuXYHYZW26} and adaptive rubric supervision~\citep{DBLP:journals/corr/abs-2511-19399}. We extend these ideas to tool-specific legal queries: downstream judgment outcomes determine trajectory preferences, from which we discover and refine rubrics for supervising intermediate retrieval actions. An extended discussion of reinforcement learning for tool-using and search agents is provided in Appendix~\ref{app:rl_related_work}.

\section{Methodology}
\label{sec:method}

\begin{figure*}[t]
    \centering
    \includegraphics[width=\textwidth]{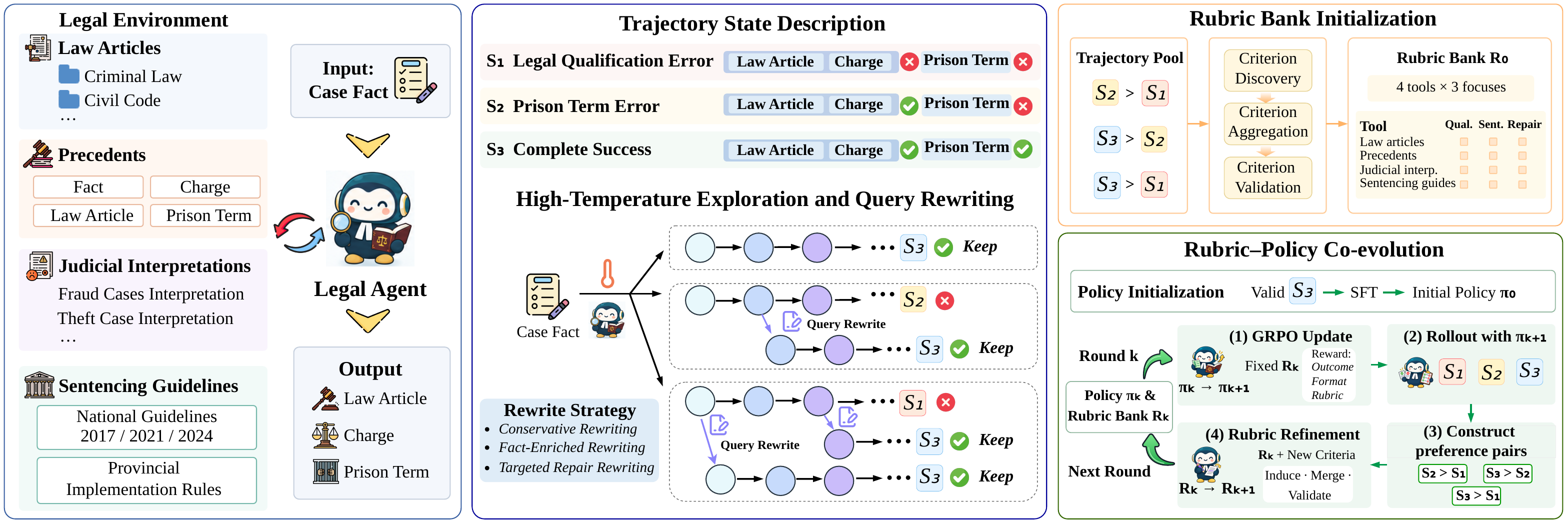}
    \caption{Task description and main pipeline of LegalEvolver.}
    \label{fig:training}
\end{figure*}

\textsc{LegalEvolver} learns legal search behaviors from diverse trajectories and updates its query evaluation criteria as the policy improves. High-temperature sampling and heuristic rewriting provide the trajectories used for policy initialization (Section~\ref{sec:trajectory_init}). \textit{Initial rubric construction} turns comparisons of these trajectories into the initial rubric bank through pairwise criterion discovery, criteria aggregation, and held-out validation (Section~\ref{sec:rubric_rl}). The bank then guides policy optimization. \textit{Rubric refinement} uses trajectories from the updated policy to discover new criteria, aggregate them with existing criteria, and revalidate the bank for the next round (Section~\ref{sec:coevolution}). Figure~\ref{fig:training} illustrates the trajectory comparisons underlying rubric construction and query scoring.

\subsection{Task Formulation}
\label{sec:task_formulation}

We first formally define the task of legal judgment prediction. Given the fact description $f$ of a case, the model is required to predict:
\begin{equation}
    y=(l,c,t),
\end{equation}
where $l$ denotes the applicable law articles, $c$ denotes the corresponding charges, and $t$ denotes the term of imprisonment. For multi-defendant cases, the output is extended to defendant-level predictions: $\{(l_i,c_i,t_i)\}_{i=1}^{n}$, where each prediction corresponds to an individual defendant.

We adopt the ReAct framework~\citep{DBLP:conf/iclr/YaoZYDSN023} as the backbone of our legal search agent. The legal agent performs multi-round interactions with external legal retrieval tools through a sequence of Reasoning--Action--Observation steps. We represent the reasoning process of the legal agent as a trajectory:
$\tau=(f,r_1,a_1,o_1,\ldots,r_m,a_m,o_m,\hat y)$,
where $r_j$ denotes the natural-language legal reasoning at step $j$, $a_j$ represents the tool-use action taken by the agent, $o_j$ denotes the corresponding retrieval observation returned by the tool, and $\hat y$ represents the final prediction generated by the agent. $\hat y$ consists of predicted law articles $\hat l$, predicted charges $\hat c$, and the predicted term of imprisonment $\hat t$.
For a retrieval action, the agent can send a query to one of the four legal knowledge bases: law articles, precedents, judicial interpretations, and sentencing guidelines. Appendix~\ref{app:environment} describes these knowledge sources and their retrieval interfaces.

To evaluate a trajectory, we compare its final prediction $\hat y=(\hat l,\hat c,\hat t)$ with the ground truth $y=(l,c,t)$ and define three outcome states:
\begin{equation}
\label{eq:trajectory_outcome}
 S(\tau)=
 \begin{cases}
 S_3, & \hat l=l,\ \hat c=c,\ \hat t=t,\\
 S_2, & \hat l=l,\ \hat c=c,\ \hat t\ne t,\\
 S_1, & \text{otherwise}.
 \end{cases}
\end{equation}
Here, $S_3$ indicates complete success, $S_2$ indicates correct legal qualification with incorrect sentencing, and $S_1$ indicates a legal qualification failure. 

\subsection{Diverse Trajectory Generation and Policy Initialization}
\label{sec:trajectory_init}

We start from a base agentic model, denoted as the initial policy $\pi_\text{base}$. Since it may repeatedly follow the same reasoning path and issue similar queries for a case, alternative ways of finding relevant evidence are left unexplored. We therefore first seek diverse trajectories for each case, increasing the opportunity to discover more useful query formulations and retrieval strategies.

\paragraph{High-temperature exploration and query rewriting.}

For each training case, we sample multiple trajectories at an elevated temperature to explore different reasoning paths and queries. Our preliminary study identifies poorly constructed queries as the main failure source, as detailed in Section~\ref{query-quality-analysis}. We analyze the failed search trajectories and group their recurring query failures into three patterns: (1)~Queries often consist only of a charge name, generic legal terms, or a specific article number, leaving the underlying legal question implicit; (2)~Queries may omit facts needed to distinguish charges or determine a sentencing range, even when the general retrieval intent is clear; (3)~Searches may remain anchored to an incorrect charge or article, conflate multiple charges, or repeat an early hypothesis without using retrieved evidence to seek new discriminating information. 

Comparing these patterns with human experts' LJP trajectories motivates three complementary rewriting strategies:
(1)~\textit{Conservative rewriting.} Human experts formulate explicit questions around the legal uncertainty they need to resolve. We therefore clarify the intended legal issue and make the query more specific while preserving its original information need.
(2)~\textit{Fact-enriched rewriting.} Human experts ground their searches in legally decisive case details. We therefore enrich queries with relevant facts from the case description, including conduct, consequences, and circumstances.
(3)~\textit{Targeted repair rewriting.} Human experts adapt their searches to unresolved issues and prior findings. We therefore use the trajectory's outcome state ($S_1$ or $S_2$), the retrieval tool, and the search history to revise the search direction: legal qualification failures ($S_1$) motivate queries about statutory elements and distinctions between candidate charges, whereas sentencing failures ($S_2$) motivate queries about sentencing thresholds and adjustment factors. Prior queries and results guide which unresolved issue to investigate next.

At a selected retrieval step, we rewrite the query while preserving the preceding history and tool. We execute the new query and regenerate the continuation to construct an enhanced trajectory: $\tau'=(f,r_1,a_1,o_1,\ldots,r_n,a_n',o_n',\ldots,r_m',a_m',o_m',\hat y')$. All natural trajectories sampled with high temperature and trajectories enhanced with rewriting enter the initial trajectory pool $P_0$.

\paragraph{Rejection sampling and SFT.} We select $S_3$ trajectories from the trajectory pool $P_0$ as the SFT training set. These trajectories yield correct final predictions, and their tool calls and outputs satisfy the required format. Then we fine-tune the initial policy $\pi_\text{base}$ on the SFT training set, supervising reasoning, tool calls, and final predictions while treating retrieved observations as context. This produces the fine-tuned policy $\pi_0$. Appendix~\ref{app:training_details} provides more details about the SFT process.

\subsection{Initial Rubric Construction and Guided Policy Optimization}
\label{sec:rubric_rl}

The initial trajectory pool $P_0$ contains both successful and failed attempts at the same cases. Although SFT learns from the successful trajectories, an outcome label alone provides little guidance on how an intermediate query could be improved. We therefore compare trajectories with different outcomes for the same case to identify query choices associated with better judgments. Such comparisons can yield a more comprehensive rubric to guide the legal search agent.

\paragraph{Initial rubric construction.}
Using the outcome states and preference ordering defined in Section~\ref{sec:task_formulation}, we construct three types of within-case preference pairs: $S_3$ versus $S_2$, $S_2$ versus $S_1$, and $S_3$ versus $S_1$, with the first trajectory preferred in each pair. These pairs provide the input to pairwise criterion discovery. These pairs are sampled from the initial trajectory pool $P_0$. Each pair of trajectories comes from the same case.
Since we have four legal retrieval bases, the legal search agent has access to 4 retrieval tools. As for the outcome dimension, we define three reasoning focuses: \textit{Sentencing} ($S_3$ versus $S_2$), \textit{Legal Qualification} ($S_2$ versus $S_1$), and \textit{Full Repair} ($S_3$ versus $S_1$). 
Thus, we organize criteria into 12 (4$\times$3) tool-focus groups, each forming a rubric with multiple criteria. Let $\mathcal U$ denote the set of four retrieval tools and $\mathcal F$ the set of three reasoning focuses. For a collection of within-case preference pairs $\mathcal P$, let $\mathcal P_{u,F}$ denote the pairs used to discover criteria for tool $u\in\mathcal U$ and focus $F\in\mathcal F$, with comparisons centered on queries issued to tool $u$.

We construct the initial rubric bank $\mathcal R_0$ in three steps. (1)~\textit{Pairwise Criterion Discovery.} We use a strong LLM to compare the preferred and rejected trajectories in the context of the case and interaction history. It identifies differences in query content and search focus, then proposes criteria for the corresponding tool-focus group. Each candidate includes a description, a four-level scoring guide, and a confidence score. (2)~\textit{Criteria Aggregation.} Within each group, we remove incomplete, ambiguous, and low-confidence candidates. Then we merge duplicates and cluster related criteria into reusable guidance. For example, observations from individual cases can yield similar criteria that require sentencing queries to include the charge, monetary amount, and mitigating circumstances. For each group, criteria aggregation combines evidence across pairs into a candidate rubric. (3)~\textit{Criteria Validation.} We sample held-out trajectory pairs $\mathcal V$ from $P_0$, with $\mathcal V_{u,F}$ denoting the pairs used to validate criteria for tool $u$ and focus $F$. We test the aggregated criteria on held-out trajectory pairs and retain those that consistently assign higher scores to preferred-side queries across sufficient examples. Finally, the retained criteria form $\mathcal R_0$. Appendix~\ref{app:rubric_details} gives the criterion representation and construction statistics.

\paragraph{Rubric-guided policy optimization.}
After constructing the initial rubric bank $\mathcal R_0$, we apply $\mathcal R_0$ to guide the RL training of our legal agent. We specifically design a rubric reward to provide fine-grained supervision for the agent. To calculate the rubric reward for a trajectory, we first identify all tool calls in the trajectory, and for each tool call, we select one of the 12 rubrics according to tool and
focus. A strong LLM serves as a rubric evaluator that scores its criteria using the query, case, and interaction history.
The typical reward for agent RL training consists of outcome reward $R_{\mathrm{outcome}}$ and format reward $R_{\mathrm{format}}$~\citep{DBLP:conf/nips/QianAHWCHTJ25}. We further add the rubric reward:
\begin{equation}
 R_{\mathrm{rubric}}(\tau)
 =\frac{1}{m}\sum_{j=1}^{m}
 \left(\sum_{c=1}^{C_j}w_{jc}\,s_{jc}\right).
 \label{eq:rubric_reward}
\end{equation}
Here $m$ is the number of retrieval tool calls in the trajectory, and $C_j$ is the number of criteria in the single rubric selected for call $j$. The score $s_{jc}$ for criterion $c$ takes one of four values: $0$, $1/3$, $2/3$, or $1$; $w_{jc}$ is its non-negative weight, with weights summing to one within the selected rubric. The inner sum computes each call's weighted rubric score; the outer average combines these scores across the trajectory. A trajectory without tool calls receives zero rubric reward.

We combine this process reward with outcome and format rewards:
\begin{equation}
 R_k(\tau)
 =\lambda_o R_{\mathrm{outcome}}(\tau)
 +\lambda_f R_{\mathrm{format}}(\tau)
 +\lambda_r R_{\mathrm{rubric}}(\tau).
\label{eq:combined_reward}
\end{equation}
The rubric term is computed using the current rubric bank $\mathcal R$; the outcome term measures law article, charge, and prison-term prediction correctness; the format term checks valid tool calls and structured final outputs (Appendix~\ref{app:training_details}). We optimize the policy with GRPO~\citep{DBLP:journals/corr/abs-2402-03300} using rewards from Equation~\ref{eq:combined_reward}. Rubric supervision supplies feedback on intermediate query construction even when the final prediction is incorrect, while outcome supervision anchors optimization to judgment accuracy. The first round starts from the SFT policy $\pi_0$ and the constructed bank $\mathcal R_0$.

\label{sec:coevolution}\begin{wrapfigure}[14]{r}{0.47\linewidth}
  \vspace{-3.7em}
\refstepcounter{algorithm}\raggedright\textbf{Algorithm \thealgorithm}\quad Rubric-Policy Co-Evolution\par\vspace{0.25em}\hrule\vspace{0.35em}\label{alg:coevolution}\small\begin{algorithmic}[1] \Require $\pi_0,\mathcal R_0,\mathcal D,\mathcal D_{\mathrm{val}},\mathcal V,\epsilon_{\mathrm{stop}}>0$ \State $k\gets0$; $v_0\gets J(\pi_0;\mathcal D_{\mathrm{val}})$ \Repeat \State $\pi_{k+1}\gets\textsc{GRPO}(\pi_k,R_k)$  \State $\mathcal T\gets\textsc{Rollout}(\pi_{k+1},\mathcal D)$ \State $\mathcal P\gets\textsc{Pair}(\mathcal T)$ \For{$(u,F)\in\mathcal U\times\mathcal F$} \State $\mathcal C\gets\textsc{Discover}(\mathcal P_{u,F})$ \State $\mathcal C\gets\textsc{Aggregate}(\mathcal R_k(u,F),\mathcal C)$ \State $\mathcal R_{k+1}(u,F)\gets\textsc{Revalidate}(\mathcal C,\mathcal V_{u,F})$ \EndFor \State $v_{k+1}\gets J(\pi_{k+1};\mathcal D_{\mathrm{val}})$ \State $\Delta\gets v_{k+1}-v_k$; $k\gets k+1$ \Until{$\Delta<\epsilon_{\mathrm{stop}}$} \State \Return $\pi_k,\mathcal R_k$\end{algorithmic}\vspace{0.2em}\hrule
\end{wrapfigure}

\subsection{Rubric-Policy Co-Evolution}

Typical agent training involves SFT followed by RL~\citep{DBLP:conf/emnlp/CaiZZSXSWB25,DBLP:journals/corr/abs-2601-07296}. To enable the agent to surpass the capability ceiling of this training paradigm, we propose a rubric-based co-evolutionary self-improvement paradigm in which the policy and rubric bank improve through alternating updates. Starting from the SFT policy $\pi_0$ and the initial rubric bank $\mathcal R_0$ constructed from trajectory contrasts, each round takes $(\pi_k,\mathcal R_k)$ as input, first updates the policy under the fixed rubric bank, and then refines the rubrics using trajectories from the updated policy. Here $k$ indexes co-evolution rounds, and $\mathcal R_k(u,F)$ denotes the rubric associated with tool $u$ and focus $F$ in the round-$k$ bank. The round ends with $(\pi_{k+1},\mathcal R_{k+1})$, which becomes the input to the next round.

\paragraph{Policy improvement with fixed rubrics.} At the start of round $k$, we take $(\pi_k,\mathcal R_k)$ as input and hold the criteria and scoring weights in $\mathcal R_k$ fixed. Each tool call is scored by its selected rubric, and the call scores are averaged as in Equation~\ref{eq:rubric_reward}. We optimize $\pi_k$ with the combined reward in Equation~\ref{eq:combined_reward}, obtaining $\pi_{k+1}$. Keeping the rubric bank fixed during policy optimization provides consistent query evaluation criteria during that policy update.

\paragraph{Rubric refinement.} In the second stage of round $k$, we use the updated policy $\pi_{k+1}$ to revise $\mathcal R_k$. As the policy's search behaviors change, the existing bank may miss newly effective strategies or emerging failure patterns. We refine the bank through three steps:
(1)~\textit{Updated-policy Trajectory Collection.} We sample trajectories from $\pi_{k+1}$ on training cases and classify their outcomes using Equation~\ref{eq:trajectory_outcome}. We construct the same three within-case preference-pair types as in initial construction, capturing query differences under the updated policy. (2)~\textit{Criterion Discovery and Aggregation.} We apply pairwise criterion discovery to the new pairs. Within each tool-focus group, we combine the resulting candidates with the existing criteria in $\mathcal R_k$, remove incomplete or low-confidence candidates, and merge and abstract overlapping criteria. Unlike initial construction, this step updates an existing rubric rather than building one from scratch. (3)~\textit{Criterion Revalidation.} We test the combined old and new criteria using the same held-out validation procedure as in initial construction. Criteria must continue to distinguish preferred-side queries to be retained, so ineffective old criteria can be removed alongside ineffective new candidates. The retained criteria form $\mathcal R_{k+1}$, completing round $k$ with $(\pi_{k+1},\mathcal R_{k+1})$ as the input to the next round.

\paragraph{Coevolution Algorithm.} Algorithm~\ref{alg:coevolution} summarizes the iteration. Here $\mathcal U$ and $\mathcal F$ denote the four tools and three focuses, $\mathcal D$ the training cases, and $\mathcal V$ the held-out trajectory contrasts. \textsc{Pair} constructs outcome-based preference pairs, \textsc{Discover} performs pairwise criterion discovery, \textsc{Aggregate} filters and aggregates old and new criteria, and \textsc{Revalidate} checks the combined criteria on $\mathcal V$. High-temperature sampling, heuristic rewriting, and rejection sampling initialize the training pool; later rounds refine the rubrics using updated-policy trajectories. On validation cases $\mathcal D_{\mathrm{val}}$, we track a fixed judgment metric $J$ (higher is better), stopping when its gain is below $\epsilon_{\mathrm{stop}}>0$.

\section{Experiments}
\label{sec:experiments}

\begin{table*}[!t]\caption{Performance comparison on the CAIL2018 and CMDL datasets. The best results are indicated in bold, and the second-best results are underlined.}\label{tab:cail2018_cmdl_performance}
  \centering
  \small
  \setlength{\tabcolsep}{3.8pt}
  \renewcommand{\arraystretch}{1.12}
  \resizebox{\textwidth}{!}{%
  \begin{tabular}{l cc cc cc cc cc cc c}
    \toprule
    \multirow{3}{*}{\textbf{Methods}}
    & \multicolumn{6}{c}{\textbf{CAIL2018}}
    & \multicolumn{6}{c}{\textbf{CMDL}}
    & \multirow{3}{*}{\textbf{Overall}} \\
    \cmidrule(lr){2-7} \cmidrule(lr){8-13}

    & \multicolumn{2}{c}{\textbf{Law Article}}
    & \multicolumn{2}{c}{\textbf{Charge}}
    & \multicolumn{2}{c}{\textbf{Prison Term}}
    & \multicolumn{2}{c}{\textbf{Law Article}}
    & \multicolumn{2}{c}{\textbf{Charge}}
    & \multicolumn{2}{c}{\textbf{Prison Term}}
    & \\
    \cmidrule(lr){2-3} \cmidrule(lr){4-5} \cmidrule(lr){6-7}
    \cmidrule(lr){8-9} \cmidrule(lr){10-11} \cmidrule(lr){12-13}

    & Acc. & Ma-F1
    & Acc. & Ma-F1
    & Acc. & Ma-F1
    & Acc. & Ma-F1
    & Acc. & Ma-F1
    & Acc. & Ma-F1
    & \\
    \midrule

    \multicolumn{14}{l}{\textit{\textbf{Classification Methods}}} \\
    TopJudge
    & 75.7 & 64.6 & 71.8 & 63.6 & 20.1 & 11.6
    & 80.4 & 82.6 & 79.5 & 80.2 & 20.2 & 11.8 & 55.2 \\
    NeurJudge
    & 74.8 & 62.0 & 71.2 & 63.0 & 18.9 & 10.6
    & 77.1 & 79.0 & 74.0 & 74.7 & 20.3 & 12.1 & 53.1 \\
    BERT
    & 80.6 & 68.3 & 76.4 & 68.7 & 19.2 & 9.8
    & 81.6 & 83.3 & 79.3 & 80.1 & 20.8 & 13.1 & 56.8 \\
    LawFormer
    & 78.7 & 65.9 & 73.3 & 64.3 & 20.5 & 12.5
    & 79.9 & 81.8 & 75.9 & 76.6 & 23.4 & 18.2 & 55.9 \\

    \midrule
    \multicolumn{14}{l}{\textit{\textbf{Direct Reasoning}}} \\
    Qwen3-8B
    & 60.9 & 42.8 & 56.9 & 34.6 & 8.7 & 6.9
    & 67.5 & 77.1 & 73.6 & 78.3 & 9.0 & 8.2 & 43.7 \\
    GLM-4-9B
    & 46.2 & 28.5 & 50.6 & 30.0 & 6.1 & 4.7
    & 66.9 & 70.8 & 62.6 & 65.1 & 6.8 & 6.2 & 37.0 \\
    Hunyuan-A13B
    & 71.8 & 52.4 & 67.1 & 42.9 & 10.3 & 8.4
    & 81.1 & 84.7 & 72.5 & 73.5 & 9.9 & 8.1 & 48.6 \\
    Qwen3-14B
    & 71.3 & 54.7 & 65.6 & 42.3 & 7.8 & 6.7
    & 73.3 & 82.7 & 69.4 & 73.0 & 10.2 & 9.5 & 47.2 \\
    GLM-4-32B
    & 57.5 & 40.5 & 60.1 & 40.1 & 3.6 & 2.9
    & 71.5 & 78.9 & 68.3 & 72.1 & 5.3 & 4.8 & 42.1 \\
    Qwen-Max
    & 78.8 & 64.5 & 76.4 & 67.9 & 10.5 & 10.7
    & 82.0 & 86.9 & \underline{83.3} & \underline{85.0} & 10.5 & 11.2 & 55.6 \\

    \midrule
    \multicolumn{14}{l}{%
      \textit{\textbf{Agentic Retrieval-Augmented Reasoning~(Qwen3-8B)}}} \\
    Agentic RAG
    & 62.9 & 44.8 & 58.2 & 35.3 & 10.2 & 8.2
    & 72.2 & 80.1 & 75.9 & 79.7 & 7.9 & 7.0 & 45.2 \\
    Search-o1
    & 63.6 & 46.3 & 67.7 & 59.0 & 7.8 & 6.7
    & 70.5 & 80.3 & 75.4 & 81.3 & 8.6 & 8.3 & 48.0 \\
    PLJP
    & 67.1 & 57.9 & 69.0 & 59.9 & 20.1 & 13.2
    & 62.0 & 64.5 & 70.9 & 71.5 & 16.4 & 8.9 & 48.5 \\
    GLARE
    & 76.9 & 63.2 & 76.1 & 68.2 & 10.3 & 9.8
    & 82.9 & 85.3 & 82.1 & 83.0 & 9.4 & 8.9 & 54.7 \\

    \midrule
    \multicolumn{14}{l}{%
      \textit{\textbf{Post-Training Methods~(Qwen3-8B)}}} \\
    Vanilla SFT
    & 79.9 & \underline{70.0} & 76.6 & 68.5 & 20.0 & 20.5
    & 80.6 & 82.7 & 79.6 & 80.2 & 23.6 & 19.7 & 58.5 \\

    Standard GRPO
    & 80.7 & 68.9 & 78.2 & 69.4 & \underline{21.2} & \underline{20.8}
    & 84.2 & 85.2 & 82.1 & 83.8 & 24.0 & 20.7 & 59.9 \\

    SmartSearch
    & \underline{81.1} & 67.3 & \underline{78.3} & \underline{69.5} & 20.7 & 20.7
    & \underline{85.2} & \underline{87.3} & 82.3 & 84.0 & \underline{25.5} & \underline{22.6} & \underline{60.4} \\

    \rowcolor{blue!10} LegalEvolver (Ours)
    & \textbf{82.1} & \textbf{70.7} & \textbf{79.4} & \textbf{71.9} & \textbf{22.1} & \textbf{21.0}
    & \textbf{85.7} & \textbf{87.6} & \textbf{84.4} & \textbf{85.1} & \textbf{25.6} & \textbf{22.8} & \textbf{61.5} \\

    \bottomrule
  \end{tabular}%
  }
  \relax
  \relax
\end{table*}

\subsection{Experimental Setup}
\paragraph{Datasets and Metrics.} 
We evaluate LegalEvolver across two complementary legal scenarios. CAIL2018~\citep{DBLP:journals/corr/abs-1807-02478} covers single-defendant cases, whereas CMDL~\citep{DBLP:conf/acl/00030LWG024} contains multi-defendant cases and therefore tests whether the method transfers to more complex fact patterns. To reduce charge-frequency bias, we uniformly sample cases across charge labels to construct balanced test sets, including 1,000 CAIL2018 cases and 2,000 CMDL cases that collectively cover 201 distinct charge categories. Detailed dataset statistics are provided in Appendix~\ref{app:dataset_details}. Following prior work, we evaluate each prediction task with Accuracy (Acc.) and Macro-F1 (Ma-F1). Full implementation and training details are provided in Appendix~\ref{app:training_details}.

\paragraph{Baselines.}
We compare our method against four categories of baseline approaches. \textbf{(1) Classification Methods:} We consider four supervised classification methods for LJP, including \textit{TopJudge}~\citep{DBLP:conf/emnlp/ZhongGTX0S18}, \textit{NeurJudge}~\citep{DBLP:conf/sigir/Yue0JWZACYW21}, \textit{BERT}~\citep{DBLP:conf/naacl/DevlinCLT19}, and \textit{LawFormer}~\citep{DBLP:journals/aiopen/XiaoHLTS21}. \textbf{(2) Direct Reasoning:} We use a range of models to predict judgments directly from case facts without external retrieval, including Qwen3-8B~\citep{DBLP:journals/corr/abs-2505-09388}, GLM-4-9B~\citep{DBLP:journals/corr/abs-2406-12793}, Hunyuan-A13B~\citep{tencenthunyuanteam2026hunyuana13btechnicalreport}, Qwen3-14B~\citep{DBLP:journals/corr/abs-2505-09388}, GLM-4-32B~\citep{DBLP:journals/corr/abs-2406-12793}, and Qwen-Max~\citep{DBLP:journals/corr/abs-2309-16609}. \textbf{(3) Agentic Retrieval-Augmented Reasoning:} \textit{Agentic RAG} combines multi-step reasoning with external legal evidence retrieval. \textit{Search-o1} interleaves reasoning and search to acquire supporting evidence~\citep{DBLP:conf/emnlp/LiDJZZZZD25}. \textit{PLJP} uses domain models to generate candidate labels and retrieve precedents for LLM prediction~\citep{DBLP:conf/emnlp/WuZL0LZS0K23}. \textit{GLARE} distinguishes confusing charges using reasoning paths retrieved from similar cases~\citep{DBLP:conf/acl/YangDD26}. \textbf{(4) Post-Training Methods:} \textit{Vanilla SFT} fine-tunes the model on selected successful trajectories. \textit{Standard GRPO} optimizes the policy using trajectory-level outcome rewards. \textit{SmartSearch} uses general principles to optimize search queries.

\subsection{Main Results}

Table~\ref{tab:cail2018_cmdl_performance} presents the main experimental results. LegalEvolver achieves the best overall performance on both datasets and outperforms all baselines. We further make the following observations.
\textbf{(1) Agentic retrieval generally improves legal judgment prediction over direct reasoning.} External retrieval provides fine-grained legal knowledge that may not be reliably available in the model's parametric knowledge, thereby grounding legal qualification and sentencing decisions in relevant evidence.
\textbf{(2) LegalEvolver consistently outperforms existing search agents through rubric-guided policy optimization.} Our rubrics guide the policy to formulate queries around unresolved legal uncertainties, enabling it to retrieve more useful evidence than retrieval access alone. In contrast, existing agents largely rely on general query optimization and often formulate overly coarse queries that omit latent legal issues and legally salient case details.
\textbf{(3) Rubric--policy co-evolution provides stronger alignment than one-stage post-training.} LegalEvolver combines outcome supervision with query-level feedback and refreshes the rubric bank using trajectories from the updated policy. The resulting improvements across all three prediction tasks indicate that iterative supervision remains aligned with the policy as its search behavior evolves. We provide controlled ablations of the key components in Appendix~\ref{app:ablation}.\begin{wrapfigure}[16]{r}{0.52\linewidth}
  \vspace{1.2em}
  \centering
  \includegraphics[width=\linewidth]{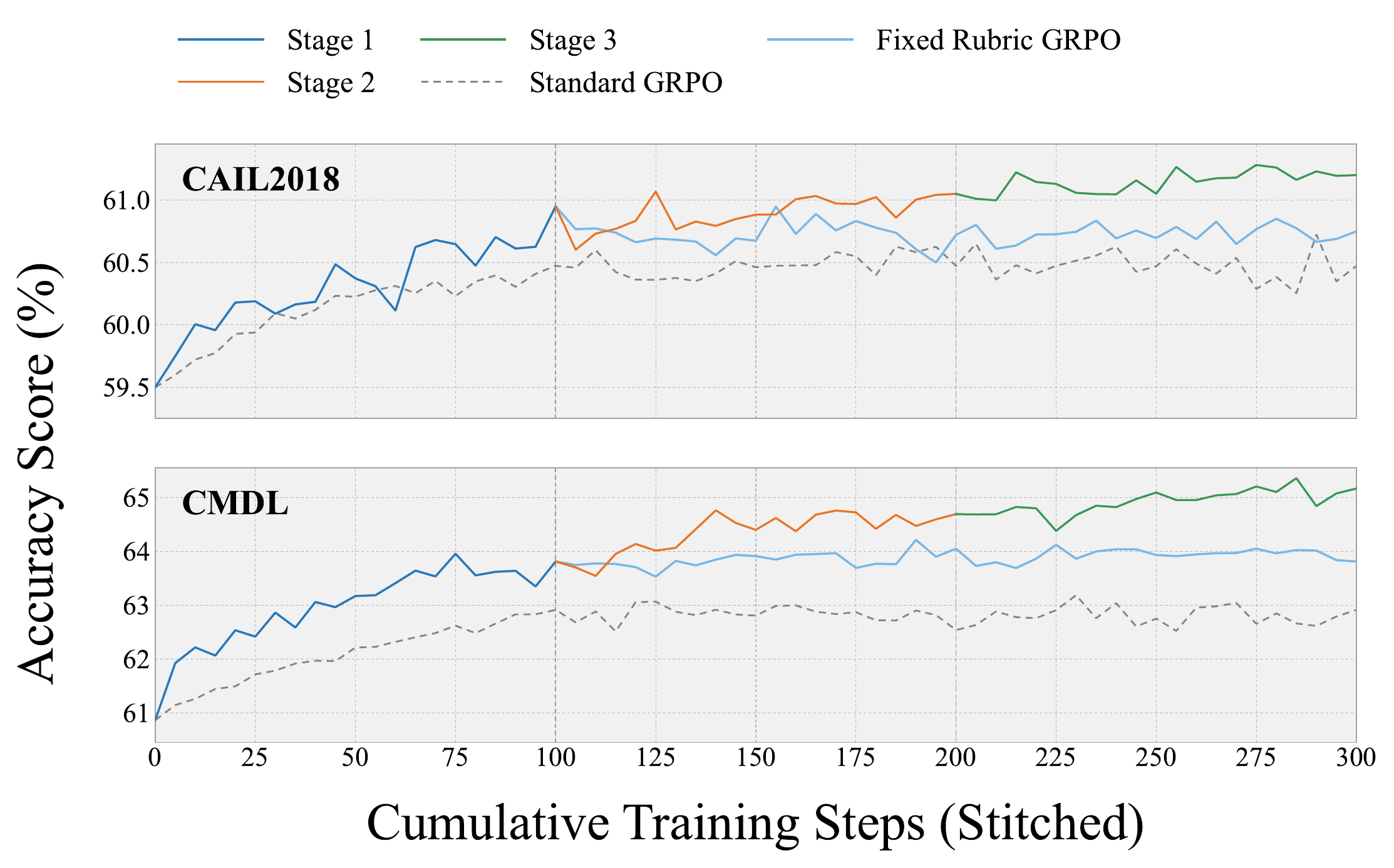}
  \vspace{-1.1em}
  \caption{Stage-wise RL training curves of LegalEvolver on CAIL2018 and CMDL.}
  \label{fig:training_dynamics}
  \vspace{-0.65em}
\end{wrapfigure}

\subsection{Quantitative Analysis}
\paragraph{Training Dynamics and Self-Evolution.}
To assess the effectiveness of query-level supervision and iterative rubric refinement, Figure~\ref{fig:training_dynamics} compares LegalEvolver with Standard GRPO and Fixed Rubric GRPO on CAIL2018 and CMDL across three training rounds. All methods start from the same SFT checkpoint and use the same total number of optimization steps. Each checkpoint reports the mean accuracy over law article, charge, and prison-term prediction. Two observations validate our design. \textbf{(1) Legal query supervision improves reward alignment.} LegalEvolver converges faster and consistently outperforms Standard GRPO, which relies only on trajectory-level outcome feedback. The steeper learning curves suggest that rewarding queries that translate case facts into targeted legal questions about specific legal uncertainties provides more effective supervision than outcome feedback alone. \textbf{(2) Rubric--policy co-evolution sustains improvement across training rounds.} Fixed Rubric GRPO uses the initial rubric bank throughout all three rounds and shows diminishing gains in later training, whereas LegalEvolver continues to improve by refining the rubric bank with trajectories generated by the updated policy. This controlled comparison rules out additional training steps as the source of improvement and shows that iterative rubric refinement keeps query supervision aligned with the policy's evolving search behavior.

\FloatBarrier
\begin{figure}[t]
  \centering
  \includegraphics[width=\textwidth]{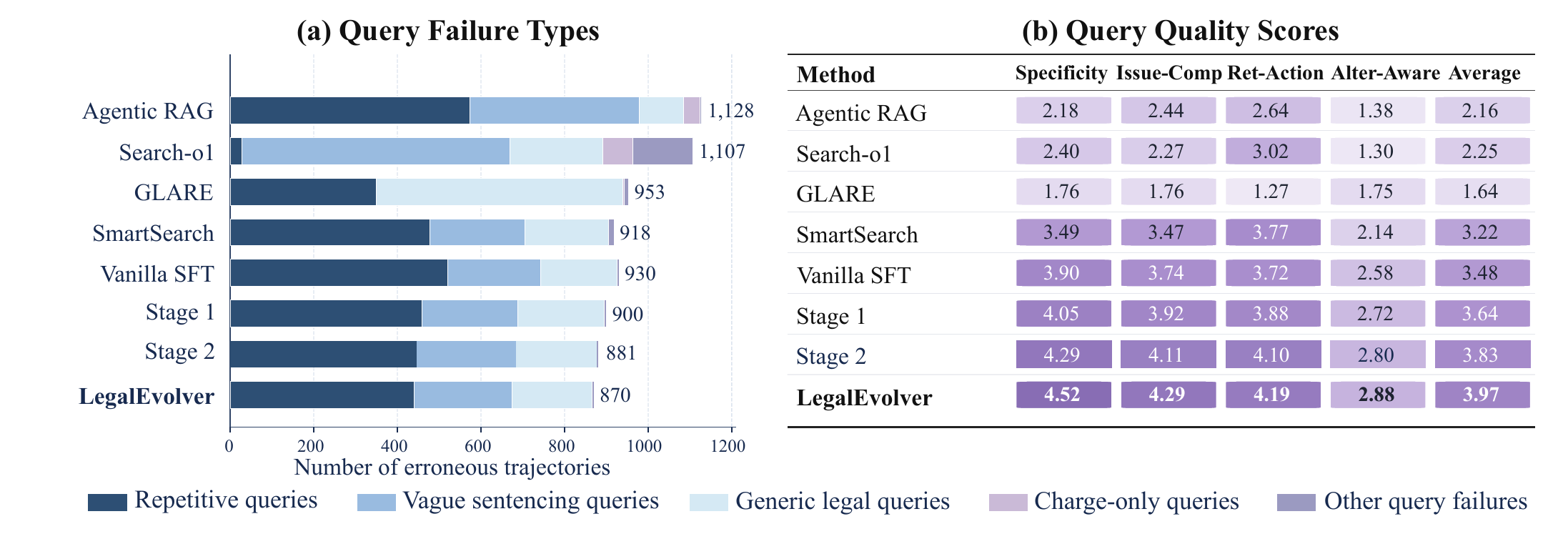}
  \caption{Query failures and quality scores across methods: (a) distribution of five query failure types and (b) mean scores on four query-quality dimensions.}
  \label{fig:query_failure_distribution}
  \label{tab:query_quality_analysis}
\end{figure}

\paragraph{Query Quality Analysis.}
\label{query-quality-analysis}
We analyze the erroneous trajectories produced by all methods and summarize their dominant query failure patterns in Figure~\ref{fig:query_failure_distribution}(a). We define a trajectory as erroneous when either its charge or law article prediction is incorrect, and use GLM-5.2~\citep{glm5team2026glm5vibecodingagentic} to assign each erroneous trajectory to an error category. The dominant failure patterns can be grouped into three broad categories: repeated searches based on incorrect assumptions, omission of key case facts, and under-specified retrieval intent. LegalEvolver produces the fewest erroneous trajectories, with fewer repetitive and coarse-grained queries. This pattern is consistent with our training strategy, which encourages the policy to formulate queries that directly address specific legal uncertainties and retrieve the evidence needed to resolve them.

To assess whether LegalEvolver addresses these failures, we evaluate query quality along four dimensions: \textit{Legal Specificity}, whether queries contain case-specific legal concepts and key facts; \textit{Legal-Issue Completeness}, whether the query set covers the main legal issues required for a complete judgment; \textit{Retrieval Actionability}, whether queries are clearly formulated and can directly guide the selected tool to retrieve useful evidence; and \textit{Alternative Awareness}, whether queries consider plausible competing legal interpretations and seek the evidence needed to distinguish among them, rather than only supporting an initial hypothesis. We use three independent LLM judges to score anonymized query sets from all methods on 300 cases from each dataset. As shown in Figure~\ref{fig:query_failure_distribution}(b), LegalEvolver ranks first on all four dimensions and achieves the highest overall score. Its advantages in \textit{Legal Specificity} and \textit{Alternative Awareness} show that the learned policy more effectively translates case facts into targeted legal questions and explores competing interpretations before reaching a judgment. In contrast, general-purpose query optimization does not fully address the task-specific information needs of legal judgment prediction.

\begin{figure*}[t]
  \centering
  \includegraphics[width=\textwidth]{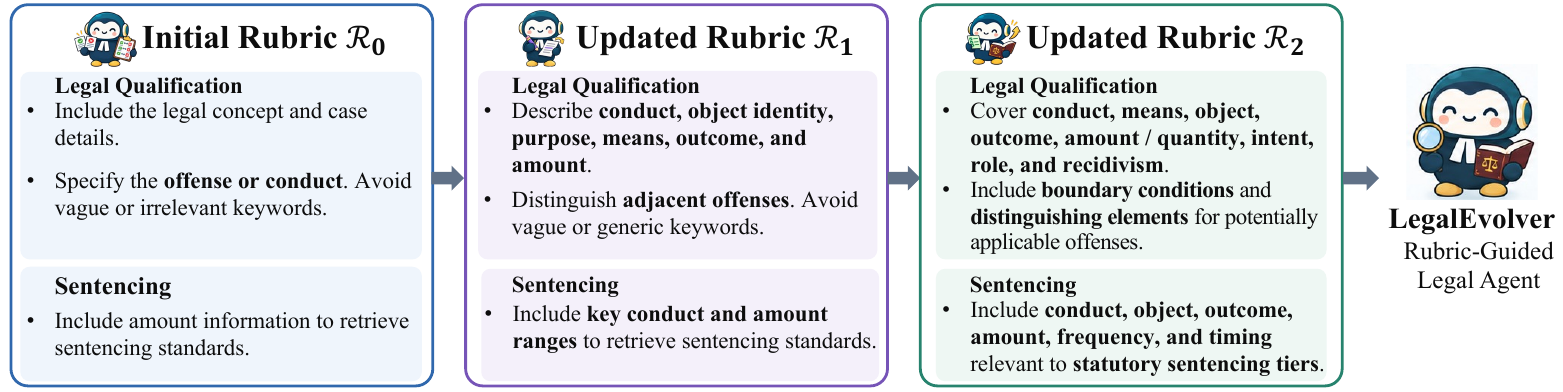}
  \caption{Evolution of the \textit{statute\_search} rubric from the initial policy through two optimization stages, shown for legal qualification and sentencing.}
  \label{fig:rubric_query_coevolution}
\end{figure*}

\paragraph{Rubric Evolution.}
Figure~\ref{fig:rubric_query_coevolution} takes the tool \textit{statute\_search} as an example to illustrate how the rubric bank evolves across training stages. For legal qualification, the rubrics progress from isolated legal concepts and case details to reusable factual dimensions and explicit criteria for identifying offense boundaries and distinguishing potentially applicable charges. For sentencing, they evolve from simple amount-related cues to structured coverage of the conduct, object, outcome, amount, frequency, and timing that determine statutory sentencing tiers. These examples show how rubric--policy co-evolution turns local retrieval heuristics revealed by updated trajectories into more comprehensive and reusable query supervision, enabling the policy to formulate searches that better address the legal uncertainty at each reasoning stage.

\section{Conclusion}
We presented \emph{LegalEvolver}, a framework that improves legal judgment prediction by explicitly optimizing the retrieval queries generated during agent reasoning. LegalEvolver induces rubrics from within-case trajectory contrasts and uses them to supervise intermediate queries during policy optimization. It then refines the rubric bank using trajectories generated by the updated policy, enabling the policy and the rubrics to co-evolve through alternating updates. Experimental results show that LegalEvolver achieves the strongest overall performance and produces more legally specific and informative queries that better address unresolved legal uncertainties. Further analyses indicate that the evolving rubrics remain aligned with changes in the policy's search behavior, enabling continued improvement beyond one-stage post-training. These findings demonstrate the value of rubric--policy co-evolution for evidence-grounded legal judgment prediction.

\clearpage

\section*{AI use statement}
In this work, we used generative AI tools to generate synthetic research data. Specifically, LLMs were used to sample and rewrite agent trajectories, induce and aggregate query rubrics from trajectory contrasts, and score intermediate queries during training. We also used AI tools to support qualitative and thematic data analysis, specifically through automated error annotation and query-quality evaluation. We did not use generative AI tools to develop theoretical models or conceptual frameworks, formulate mathematical claims, provide critical ingredients for proving mathematical claims, assist in writing proofs, propose or refine hypotheses, design or provide feedback on the research methodology or experiments, implement methods, assist with translation, clean or reformat datasets, or interpret results. All AI-assisted data, rubrics, annotations, evaluations, and analyses included in this work were reviewed and verified by the authors. We take responsibility for the final content of this work, including any text, claims, or artifacts produced with the assistance of generative AI.

\section*{Ethics statement}
This work studies legal judgment prediction using the publicly available CAIL2018 and CMDL research datasets together with public legal materials, including statutes, judicial interpretations, sentencing guidelines, and precedents. We do not collect new personal data or attempt to identify individuals, and we report results in aggregate without reproducing identifying case details. Nevertheless, criminal cases may contain sensitive facts, and automated legal predictions can inherit errors and social biases from the underlying data, retrieval corpora, and language models. The resulting system is intended solely for research and should not be used as legal advice or as a substitute for qualified legal professionals and judicial decision-makers. Any practical deployment would require jurisdiction-specific validation, appropriate privacy safeguards, careful bias assessment, and meaningful human oversight.

\section*{Reproducibility statement}
We provide detailed descriptions of the method and experimental protocol to facilitate reproducibility. The main text specifies the task formulation, retrieval environment, rubric--policy co-evolution procedure, baselines, evaluation metrics, and inference budget. The appendix further documents dataset statistics, the four retrieval tools, training-data construction, rubric induction and validation, reward definitions, and the complete SFT and GRPO configurations. Agentic baselines use the same retrieval tools and maximum number of tool calls, and controlled GRPO variants use the same initialization, training data, and optimization budget.

\clearpage
\bibliographystyle{iclr2027_conference}
\bibliography{references}

\clearpage
\appendix\section{Additional Related Work}
\label{app:rl_related_work}

\paragraph{Reinforcement Learning for LLM Agents.}
Reinforcement learning has advanced tool-augmented reasoning through self-taught tool use in START~\citep{DBLP:conf/emnlp/LiXZYZYHLWL25} and reward-driven tool learning in ToolRL~\citep{DBLP:conf/nips/QianAHWCHTJ25}. ReSearch applies RL to interleaved reasoning and search~\citep{DBLP:conf/nips/ChenSLSZZWPZCYZ25}. However, outcome-only rewards provide sparse feedback for intermediate search decisions, motivating step-wise supervision~\citep{DBLP:conf/emnlp/ZhengAWWW25}. In the legal domain, MSR$^2$ targets sentencing-factor identification~\citep{DBLP:journals/corr/abs-2602-04690}, while SyLeR uses structure-aware rewards to encourage syllogistic reasoning~\citep{DBLP:conf/cikm/Zhang0S025}. In general-purpose search, SmartSearch evaluates query novelty, intent necessity, and retrieval relevance~\citep{DBLP:conf/sigir/WenDD26}, while StepSearch uses information gain and redundancy penalties for step-wise RL~\citep{DBLP:conf/emnlp/ZhengAWWW25}. These methods motivate our complementary focus on intermediate query construction and legally grounded search feedback.

\paragraph{Innovations beyond existing methods.}
Beyond its application to legal retrieval, LegalEvolver makes two methodological advances. \textbf{(1) Compared with existing legal search agents,} it converts within-case contrasts between successful and failed trajectories into explicit supervision for intermediate query construction. The resulting rubrics directly reward whether a query addresses the unresolved legal issue, rather than relying only on final judgment accuracy or general reasoning structure. \textbf{(2) Compared with general-domain evolving-rubric methods such as DR Tulu~\citep{DBLP:journals/corr/abs-2511-19399},} LegalEvolver conditions the rubric bank on both retrieval tool and reasoning focus, so different knowledge sources and reasoning stages receive distinct criteria. The bank is then updated from the improved policy's trajectories and filtered through held-out revalidation, coupling rubric evolution to fine-grained retrieval decisions.

\section{Dataset Details}
\label{app:dataset_details}

\begin{figure*}[t]
    \centering
    \begin{minipage}[t]{0.43\textwidth}
        \centering
        \textbf{(a) Test-set Statistics}\par\vspace{0.6em}
        \footnotesize
        \setlength{\tabcolsep}{4pt}
        \renewcommand{\arraystretch}{1.32}
        \begin{tabular}{@{}lcc@{}}
            \toprule
            \textbf{Dataset} & \textbf{CAIL2018} & \textbf{CMDL} \\
            \midrule
            \# Test cases & 1,000 & 2,000 \\
            \# Charges & 189 & 117 \\
            \# Articles & 140 & 106 \\
            Defendants per case & 1 & 2--28 \\
            Avg. defendants per case & 1 & 2.95 \\
            Avg. case length & 438.3 & 1,226.13 \\
            \bottomrule
        \end{tabular}
    \end{minipage}
    \hfill
    \begin{minipage}[t]{0.54\textwidth}
        \centering
        \textbf{(b) Charge Coverage by Category}\par\vspace{0.1em}
        \includegraphics[width=0.98\linewidth]{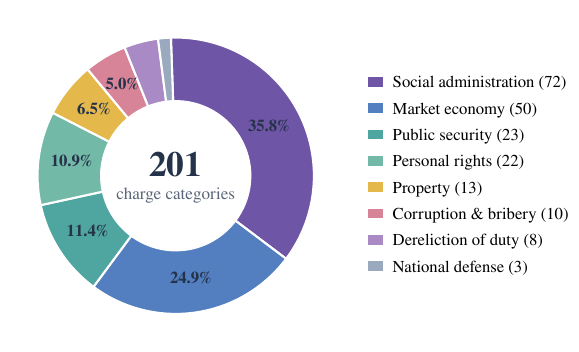}
    \end{minipage}
    \caption{Test-set statistics and the distribution of the 201 normalized charge categories covered by the two datasets.}
    \label{fig:dataset_overview}
\end{figure*}

Figure~\ref{fig:dataset_overview} summarizes the two balanced test sets and their charge coverage. Together, the two test sets cover 201 distinct charges. We group these charges into eight categories according to the taxonomy of the Criminal Law of China. This broad coverage reflects the range of charge categories encountered in real-world legal cases.

\section{Multi-source Legal Retrieval Environment}
\label{app:environment}

Legal judgment prediction requires dedicated reasoning over fine-grained legal knowledge, rather than relying solely on the knowledge stored in model parameters. To support agentic legal reasoning with reliable external evidence, we construct a multi-source legal knowledge environment consisting of law articles, precedents, judicial interpretations, and sentencing guidelines, together with their corresponding retrieval tools. These four knowledge sources provide complementary evidence from different perspectives, namely statutory constituent elements, judicial experience, rules for legal application, and sentencing standards, thereby supporting article prediction, charge prediction, and sentence prediction.

\paragraph{Law article retrieval.}
This tool retrieves relevant criminal law provisions from a database containing 491 articles. Each entry combines the article number with its statutory text as the dense-search representation. At retrieval time, semantic similarity is augmented with source-specific lexical signals for article numbers, charge names, and legally salient case terms before the final evidence is returned to the agent.

\paragraph{Precedent retrieval.}
This tool retrieves factually similar cases from China Judgments Online. The factual description is encoded for dense search, while the returned evidence preserves the corresponding law articles, charges, prison term, and available sentencing factors. Source-specific reranking further emphasizes overlap in charges, statutory references, amounts, and salient circumstances.

\paragraph{Judicial interpretation retrieval.}
This tool searches 7,214 clause-level judicial-interpretation entries. The dense representation concatenates the interpretation title, clause text, and related criminal-law articles. Reranking combines semantic similarity with matches on related articles, charges, monetary thresholds, temporal cues, and other legally salient terms.

\paragraph{Sentencing guideline retrieval.}
This tool retrieves 64 sections from the national sentencing-guideline index used during training and evaluation. Each section is represented by its source title, heading, and guideline text. Reranking favors matches on charges, amounts, aggravating or mitigating circumstances, and temporal cues, and prioritizes the guideline scope appropriate to the CAIL2018 cases.

Across all four tools, we encode retrieval text with the 768-dimensional SAILER-zh encoder~\citep{DBLP:conf/sigir/LiACDW0CT23}, L2-normalize the embeddings, and use exact FAISS inner-product search. Queries are restricted to 4--160 characters and encoded with a maximum length of 256 tokens on CPU. Each call retrieves 24 dense candidates, applies source-specific legal reranking, and returns at most four \textit{evidence cards}; returned snippets are truncated to 520 characters for statutes and 620 characters for the other sources.

\section{Rubric Construction Details}
\label{app:rubric_details}

\paragraph{Initial contrast statistics.}
The initial pool yields 9,143 natural preference pairs: 1,394 legal qualification pairs, 5,285 sentencing pairs, and 2,464 full-repair pairs. Comparing original trajectories with improved rewritten continuations yields another 11,701 pairs, for 20,844 pairs in total. Natural pairs retain the complete trajectories on both sides; rewritten pairs additionally record the original and modified queries at the branching step.

\paragraph{Rubric representation and aggregation.}
Each candidate criterion is represented as $r=(u,F,c,g,\gamma)$, where $u$ is the retrieval tool, $F$ the reasoning focus, $c$ the natural-language criterion, $g$ a scoring guide with scores in $\{0,1,2,3\}$, and $\gamma\in[0,1]$ the induction confidence. The bank contains four tool groups, each divided into Sentencing ($S_3$ versus $S_2$), Legal Qualification ($S_2$ versus $S_1$), and Full Repair ($S_3$ versus $S_1$), giving 12 rubrics in total. Filtering, clustering, and abstraction produce 340 candidate criteria; the final bank retains criteria across 10 populated tool--focus groups in $\mathcal R_0$, while two of the 12 possible combinations contain no retained criterion. These criterion counts describe initialization; subsequent rounds refine the contents of the rubrics while preserving the tool-focus hierarchy. We use Qwen2.5-72B-Instruct~\citep{DBLP:journals/corr/abs-2412-15115} as the rubric induction model to summarize the query differences between preferred and rejected trajectories into candidate criteria and aggregate related criteria within each tool-focus group.

\paragraph{Validation.}
Candidate criteria are evaluated on held-out preference pairs. Queries from natural trajectory pairs are compared only when they use the same retrieval tool; original--rewritten pairs supply a direct query correspondence at the branching step. A criterion is retained when preferred-side queries consistently score higher across sufficient validation examples. This evaluates the criterion's ability to distinguish observed search behaviors.

\section{Experiment Settings and Reward Details}
\label{app:training_details}

\paragraph{Training data construction.}
We construct SFT and GRPO data from the training portions of CAIL2018 and CMDL. During SFT data construction, each complete agent trajectory is decomposed into step-level examples for first-step planning from the case facts, next-action prediction under tool observations, and final legal-judgment prediction. This procedure yields 19,111 step-level SFT examples. The GRPO training set contains 1,448 cases drawn from the two datasets. For each GRPO round, we repeatedly sample with the model obtained from the preceding training round and retain cases that remain difficult across multiple attempts.

\paragraph{Training configuration.}
We fine-tune Qwen3-8B~\citep{DBLP:journals/corr/abs-2505-09388} with LoRA using LLaMA-Factory, retaining retrieved observations in the input context while applying the loss only to the next assistant response. Reinforcement learning is initialized from the merged SFT checkpoint and optimized with GRPO in VERL, using SGLang for multi-turn rollouts and PyTorch SDPA for policy optimization. For policy optimization, we sample eight rollouts per prompt, with up to 8,192 generation tokens, a 20,480-token context, and at most 10 retrieval calls per rollout. Tables~\ref{tab:sft_hyperparameters} and~\ref{tab:rl_hyperparameters} summarize the corresponding settings. All GRPO rounds use the same configuration reported in Table~\ref{tab:rl_hyperparameters}. For a controlled comparison, all agentic baselines use the same four legal retrieval tools as LegalEvolver and are limited to at most 10 tool calls per case.

\begin{table*}[t]
\centering
\begin{minipage}[t]{0.48\textwidth}
\vspace{0pt}
\centering
\caption{Key hyperparameters in the SFT phase.}
\label{tab:sft_hyperparameters}
\footnotesize
\setlength{\tabcolsep}{3.5pt}
\renewcommand{\arraystretch}{0.98}
\begin{tabular}{@{}p{0.60\linewidth}p{0.32\linewidth}@{}}
\toprule
\textbf{Hyperparameter} & \textbf{Value} \\
\midrule

LoRA rank / alpha / dropout & 16 / 32 / 0.05 \\

Maximum sequence length & 12,288 \\
Training epochs & 2 \\
Effective global batch size & 32 \\

Learning rate & $8\times10^{-5}$ \\
Scheduler / warm-up ratio & Cosine / 0.03 \\Precision & bfloat16 \\Number of GPUs & 8 \\

\bottomrule
\end{tabular}
\end{minipage}
\hfill
\begin{minipage}[t]{0.48\textwidth}
\vspace{0pt}
\centering
\caption{Key hyperparameters in the RL phase.}
\label{tab:rl_hyperparameters}
\footnotesize
\setlength{\tabcolsep}{3.5pt}
\renewcommand{\arraystretch}{0.98}
\begin{tabular}{@{}p{0.60\linewidth}p{0.32\linewidth}@{}}
\toprule
\textbf{Hyperparameter} & \textbf{Value} \\
\midrule

Train prompt batch size & 8 \\
Rollouts per prompt & 8 \\

Learning rate & $1\times10^{-7}$ \\
KL coefficient / type & 0.02 / low-var \\

Sampling ($T$ / top-$p$ / top-$k$) & 0.8 / 0.95 / 20 \\
Maximum response length & 8,192 \\
Maximum model length & 20,480 \\
Maximum tool calls & 10 \\

\bottomrule
\end{tabular}
\end{minipage}
\end{table*}

\paragraph{Supervised initialization.}
Retained trajectories are decomposed into next-step prediction examples $\mathcal D_{\mathrm{step}}=\{(f,h_j,b_j)\}$, where $b_j$ denotes the next agent-generated response, including reasoning and a tool call or final prediction. Retrieved observations are included in $h_j$ but are not supervised targets. The objective is
\begin{equation}
 \mathcal L_{\mathrm{SFT}}
 =-\sum_{(f,h_j,b_j)\in\mathcal D_{\mathrm{step}}}
 \log\pi_\theta(b_j\mid f,h_j).\end{equation}

\paragraph{Reward details.}
The outcome reward combines correctness scores for applicable law articles, charges, and prison terms:
\begin{equation}
 R_{\mathrm{outcome}}(\tau)
 =\alpha_l S_l+\alpha_c S_c+\alpha_t S_t,
\end{equation}
where $S_l$, $S_c$, and $S_t$ denote the corresponding prediction scores and the $\alpha$ coefficients control their contributions.
The format term checks tool-call schema validity and final-output validity:
\begin{equation}
 R_{\mathrm{format}}(\tau)
 =\tfrac{1}{2}\bigl(I_{\mathrm{tool}}(\tau)+I_{\mathrm{output}}(\tau)\bigr).
\end{equation}
The binary indicator $I_{\mathrm{tool}}$ checks valid tool names and required parameters, and $I_{\mathrm{output}}$ checks the required structured answer format. For the rubric reward, Qwen2.5-72B-Instruct scores each query using the rubric bank associated with the invoked retrieval tool: given the tool-call context, the model first identifies the current reasoning focus, selects the corresponding rubric, and then assigns the query score.

\paragraph{Query quality evaluation.}
For the query quality analysis in Figure~\ref{fig:query_failure_distribution}(b), we use three independent LLM judges: GLM-5.2, GPT-4o-mini~\citep{DBLP:journals/corr/abs-2410-21276}, and DeepSeek-v4-Flash~\citep{DBLP:journals/corr/abs-2606-19348}. Each judge anonymously evaluates the query sets along the four dimensions defined in the main text, and we average their scores to obtain the reported results.

\section{Ablation Study}
\label{app:ablation}

\paragraph{Ablation settings.}
Table~\ref{tab:ablation} isolates the contribution of each component. All GRPO-based variants use the same training data, optimization steps, and training configuration as LegalEvolver. \emph{w/o rubric evolution} keeps the initial rubric bank fixed and does not update it using trajectories generated by the improved policy. \emph{w/o rubric rewards} retains only the outcome and format rewards. \emph{w/o targeted rubrics} replaces the rubric bank organized by retrieval tool and reasoning focus with a single shared rubric for all queries. \emph{w/o GRPO} removes reinforcement learning, whereas \emph{w/o SFT} uses the base model without subsequent post-training.

\makeatletter\setlength{\@fptop}{0pt}\setlength{\@dblfptop}{0pt}\makeatother\begin{table*}[!t]
\centering
\caption{Ablation results on CAIL2018 and CMDL. Acc. and Ma-F1 are averaged over law-article, charge, and prison-term prediction within each dataset. Parentheses show percentage-point changes from LegalEvolver; higher is better for all metrics. The best results are shown in bold.}
\label{tab:ablation}
\setlength{\tabcolsep}{4pt}
\renewcommand{\arraystretch}{1.08}
\small
\begin{tabular*}{\textwidth}{@{}l@{\extracolsep{\fill}}ccccc@{}}
\toprule
\multirow{2}{*}{Method} & \multicolumn{2}{c}{CAIL2018} & \multicolumn{2}{c}{CMDL} & \multirow{2}{*}{Overall} \\
\cmidrule(lr){2-3}\cmidrule(lr){4-5}
& Acc. & Ma-F1 & Acc. & Ma-F1 & \\
\midrule
\textbf{LegalEvolver (Ours)} & \textbf{61.20} & \textbf{54.53} & \textbf{65.23} & \textbf{65.17} & \textbf{61.53} \\
w/o Rubric Evolution & 60.03 {\color{red}\scriptsize(-1.17)} & 52.50 {\color{red}\scriptsize(-2.03)} & 64.33 {\color{red}\scriptsize(-0.90)} & 64.63 {\color{red}\scriptsize(-0.54)} & 60.38 {\color{red}\scriptsize(-1.15)} \\
w/o Rubric Rewards & 60.03 {\color{red}\scriptsize(-1.17)} & 53.03 {\color{red}\scriptsize(-1.50)} & 63.43 {\color{red}\scriptsize(-1.80)} & 63.23 {\color{red}\scriptsize(-1.94)} & 59.93 {\color{red}\scriptsize(-1.60)} \\
w/o Targeted Rubrics & 60.01 {\color{red}\scriptsize(-1.19)} & 52.77 {\color{red}\scriptsize(-1.76)} & 63.88 {\color{red}\scriptsize(-1.35)} & 63.93 {\color{red}\scriptsize(-1.24)} & 60.15 {\color{red}\scriptsize(-1.38)} \\
w/o GRPO & 58.83 {\color{red}\scriptsize(-2.37)} & 53.00 {\color{red}\scriptsize(-1.53)} & 61.27 {\color{red}\scriptsize(-3.96)} & 60.87 {\color{red}\scriptsize(-4.30)} & 58.49 {\color{red}\scriptsize(-3.04)} \\
w/o SFT & 48.23 {\color{red}\scriptsize(-12.97)} & 34.57 {\color{red}\scriptsize(-19.96)} & 50.97 {\color{red}\scriptsize(-14.26)} & 55.07 {\color{red}\scriptsize(-10.10)} & 47.21 {\color{red}\scriptsize(-14.32)} \\
\bottomrule
\end{tabular*}
\end{table*}

\paragraph{Ablation analysis.}
LegalEvolver achieves the best prediction performance across all metrics. We make three observations. \textbf{(1) Continual rubric evolution improves the policy.} Keeping the initial rubric bank fixed consistently reduces performance, showing that updating the rubrics with trajectories from the improved policy allows the supervision to adapt to changes in its search behavior and provide further gains. \textbf{(2) Rubric rewards, rather than additional optimization steps, drive the improvement.} Without rubric rewards, outcome-only GRPO is trained on the same data for the same total number of steps, yet it still falls clearly behind LegalEvolver. This controlled comparison rules out extra training steps as the source of the gain and demonstrates the value of direct query-level supervision. \textbf{(3) Rubric specialization is important.} A single generic rubric provides limited improvement, whereas organizing the rubric bank by retrieval tool and reasoning focus yields consistently better performance. This result indicates that effective query criteria depend on both the knowledge source being searched and the legal issue being resolved.

\end{document}